\documentclass[twocolumn,10pt,cleanfoot,cleanheader]{asme2ej}

\usepackage{psfrag}
\usepackage{mathtools}
\usepackage{graphicx}
\usepackage{amsmath}
\usepackage{amssymb}
\usepackage{subfig}
\usepackage{environ}
\usepackage{color}

\usepackage[colorlinks,driverfallback=dvips,bookmarks=true,pagebackref,breaklinks=true]{hyperref}
\hypersetup{
	pdfauthor={Marceau MÉTILLON, Philippe CARDOU, Kévin SUBRIN, Camilo CHARRON, Stéphane CARO},
	pdftitle={A Cable-Driven Parallel Robot with Full-Circle End-Effector Rotations},
	pdfsubject={A Cable-Driven Parallel Robot with Full-Circle End-Effector Rotations}
}
\usepackage{cleveref}

\title{A Cable-Driven Parallel Robot with Full-Circle End-Effector Rotations}

\author{Marceau M\'etillon\textsuperscript{1,2},
	{\tensfb Philippe Cardou\textsuperscript{3},}
	{\tensfb K\'evin Subrin\textsuperscript{1,4},}
	{\tensfb Camilo Charron\textsuperscript{1,5},}
	{\tensfb St\'ephane Caro\textsuperscript{1,2}\thanks{Address all correspondence to this author.}}
	\affiliation{
		\textsuperscript{1}~Laboratoire des Sciences du Num\'erique de Nantes (LS2N), UMR CNRS 6004, 44300 Nantes, France\\
		\textsuperscript{2}~Centre National de la Recherche Scientifique (CNRS), 44321 Nantes, France\\
		\textsuperscript{3}~Laboratoire de robotique, D\'epartement de g\'enie m\'ecanique, Universit\'e Laval, Qu\'ebec, QC, Canada\\
		\textsuperscript{4}~Univ\'ersit\'e de Nantes, IUT, 44470 Carquefou, France\\
		\textsuperscript{5}~\'Ecole Centrale de Nantes, 44321 Nantes, France\\
		Emails: marceau.metillon@ls2n.fr, philippe.cardou@gmc.ulaval.ca,\\ kevin.subrin@ls2n.fr, camilo.charron@ls2n.fr, stephane.caro@ls2n.fr}
}

\begin{document}
	
	\maketitle    
	
	
	\begin{abstract}
		{\it Cable-Driven Parallel Robots (CDPRs) offer high payload capacities, large translational workspace and high dynamic performances. The rigid base frame of the CDPR is connected in parallel to the moving platform using cables. However, their orientation workspace is usually limited due to cable/cable and cable/moving platform collisions. This paper deals with the design, modelling and prototyping of a hybrid robot. This robot, which is composed of a CDPR mounted in series with a Parallel Spherical Wrist (PSW), has both a large translational workspace and an unlimited orientation workspace. It should be noted that the six degrees of freedom~(DOF) motions of the moving platform of the CDPR, namely, the base of the PSW, and the three-DOF motion of the PSW are actuated by means of eight actuators fixed to the base. As a consequence, the overall system is underactuated and its total mass and inertia in motion is reduced.}
	\end{abstract}
	
	\section{Introduction}\label{sec:introduction}
	
	A Cable-Driven Parallel Robot (CDPR) belongs to a particular class of parallel robots where a moving platform is linked to a base frame using cables. Motors are mounted on a rigid base frame and drive winches. Cable coiled on these winches are routed through exit points located on the rigid frame to anchor points on the moving platform. The pose (position and orientation) of the moving platform is determined by controlling the cable lengths.\par
	
	CDPRs have several advantages compared to classical parallel robots. They are inexpensive and can cover large workspaces~\cite{Hussein.18}. The lightweight cables contribute to the lower inertia of the moving platform and consequently to a better dynamic performance over classical parallel robots~\cite{Kawamura.95}. Another characteristic of the CDPRs is their reconfigurability. Changing the overall geometry of the robot can be done by changing the exit points and anchor points.\par
	
	Reconfigurability of the CDPRs is suitable for versatile applications especially in an industrial context~\cite{Gagliardini.16,Rasheed.20}. CDPRs have drawn researchers' interests towards robotic applications such as pick-and-place operations, robotic machining, manipulation, intralogistics measurements and calibration systems~\cite{Qian.18,Picard.18}.\par
	
	\begin{figure*}[!htbp]
				\centering				
				\includegraphics[width=1\linewidth]{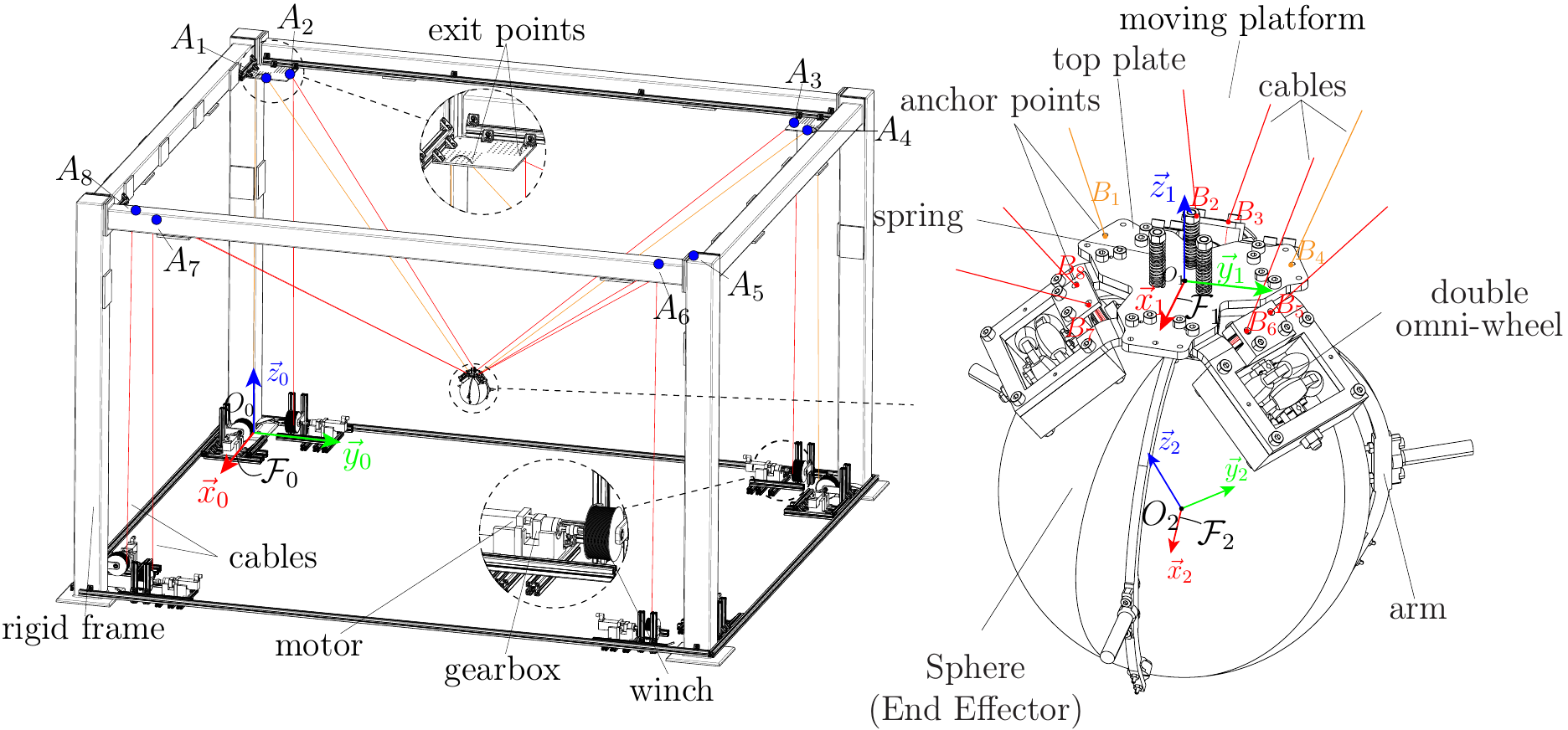}
				\caption[Manipulator]{CDPR with full-circle end-effector rotations}
				\label{fig:PSW2_CREATOR_detail}
	\end{figure*}
	
	CDPRs can offer an extremely large three-degrees of freedom translational workspace, but their orientation workspace is usually limited due to cable/cable and cable/moving platform collisions\cite{Blanchet.14}. Cable interferences can be divided into cable-cable  and cable-environment interferences~\cite{Bak.19}. For a given pose of the moving platform, cable-cable collision may occur by changing the moving platform orientation. Cable-environment collisions refer to the interferences between the moving platform and its surrounding environment.\par
	
	This paper presents the concept of a hybrid manipulator with decoupled translation and orientation motions of the moving platform. It is called hybrid since two parallel mechanisms are connected in series~\cite{Merlet.06}, namely, a CDPR and a Parallel Spherical Wrist~(PSW). In \cite{Miermeister.14}, a hybrid CDPR equipped with multiple platforms and up to nine cables is able to orientate a end-effector around an axis with an unlimited range of rotation. In~\cite{Nagai.11, Nagai.12}, redundant drive wire mechanisms are used for producing motions with high acceleration and good precision. In~\cite{Khakpour.14, Khakpour.14b, Khakpour.15}, differential cables are used to increase the size of the manipulator workspace. Differential cables consists into a set of two independent cables connecting the platform to two different winches, the latter being actuated by a differential mechanism driven by a single actuator. In~\cite{Liu.12}, two spring-loaded cable-loops allows for the control of a two-DOF planar CDPR with only two actuators. Loaded springs ensure the compliance given by the variation of cable lengths and grants a higher stiffness to the mechanism. In~\cite{VikranthReddy.19}, a bi-actuated cable is used in order to improve the orientation capacity of the end-effector of planar cable-driven robots. In~\cite{Lessanibahri.18} a hybrid CDPR uses a cable-loop for remote actuation of an embedded hoist mechanism on the moving platform. In~\cite{Lessanibahri.19}, a moving-platform embedding a two-DOF differential gear set mechanism is presented allowing the two-DOF unlimited rotational motions of the end-effector. Thanks to cable-loops (bi-actuated cables), the embedded mechanisms are actuated by transmitting power through cables from motors, which are fixed on the ground, to the moving platform~\cite{Le.16}. Therefore, by remote actuation through cable-loops, the tethering of the power cable to the moving platform is eliminated~\cite{Lessanibahri.20}. Moreover, a lower mass and a lower inertia of the moving platform are obtained due to the remote actuators.\par
	
	In this paper the mechanical design of the manipulator under study is presented in Section~\ref{sec:DESCRIPTION}. Its kineto-static model is described in Section~\ref{sec:KINETOSTATIC}. The static workspace of the manipulator is studied in Section~\ref{sec:WORKSPACE}. The methodology followed to determine the optimal cable arrangement of the system is explained in Section~\ref{sec:ARRANGEMENT}. The developed prototype of the mechanism at hand and some experimental results are shown in Section~\ref{sec:DEMONSTRATOR}. Conclusions and future work are drawn in Section~\ref{sec:CONCLUSION}.\par
	
	\section{Description and Parametrization of the Manipulator}\label{sec:DESCRIPTION}
	
	The kinematic architecture of the manipulator consists of two parallel manipulators mounted in series. A CDPR grants a large translation workspace and a PSW grants an unlimited orientation workspace. This hybrid manipulator is able to combine advantages of both mechanisms in terms of large translation and orientation workspaces.\par
	
	\subsection{Overall Architecture}\label{subsec:DESCRIPTION_overall_architecture}
	
	Figure~\ref{fig:PSW2_CREATOR_detail} shows the overall architecture of the manipulator with its main components, namely, winches, exit-points and the moving platform. The winches control the cable lengths, which move and actuate the moving platform. Cables are routed through exit-points located on the rigid frame and connected to anchor-points located on the moving platform.\par
	
	Figure~\ref{fig:PSW2_CREATOR_detail} shows the moving platform, which hosts a top plate assembly and embeds the PSW. The end-effector of the wrist is a sphere actuated by three cable-loops, which transmit the required power from motors fixed on the ground to the end-effector of the moving platform, namely the sphere.\par 
	
	The PSW is described in detail in Sec.~\ref{subsec:DESCRIPTION_3DOF_PSW} while the cable-loop system is presented in Sec.~\ref{subsec:DESCRIPTION_cable_loop}. The top plate has the six-DOF and the PSW grants a large orientation workspace to the sphere providing an overall nine-DOF workspace to the spherical end-effector regarding the base frame $\mathcal{F}_0$.\par
	
	\subsection{Three-DOF Parallel Spherical Wrist}\label{subsec:DESCRIPTION_3DOF_PSW}
	
	The concept of the PSW relies on the Atlas platform principle~\cite{Hayes.05}. An end-effector is linked to the base of the wrist using a spherical joint. Three omni-wheels are linked to the wrist base with revolute joints. The end-effector is a sphere actuated by the rotation of the omni-wheels. The position of the wheel relative to the sphere surface was defined to allow a singular-free and unlimited orientation of the sphere around the global $\vec{x}_0$, $\vec{y}_0$ and $\vec{z}_0$ axes as discussed in~\cite{Platis.17}.\par
	
	Here, the top plate of the CDPR amounts to the wrist base. Three carriage sub-assemblies are rigidly attached to the top plate. Every carriage hosts an omni-wheel. A three rigid arm platform hosts the sphere using three caster balls. Three anti-backlash compression springs are mounted on threaded rods using nuts, thus ensuring the connection of the arm platform to the top plate. The springs ensure an adjustable contact force of the omni-wheels on the sphere. The omni-wheels transmit torque to the sphere thanks to friction. Each omni-wheel is independently driven by a cable-loop system.\par
		
	\subsection{Cable-Loop Principle}\label{subsec:DESCRIPTION_cable_loop}
	
\begin{figure}[!htbp] 
		\begin{center}
			\includegraphics[width=1\linewidth]{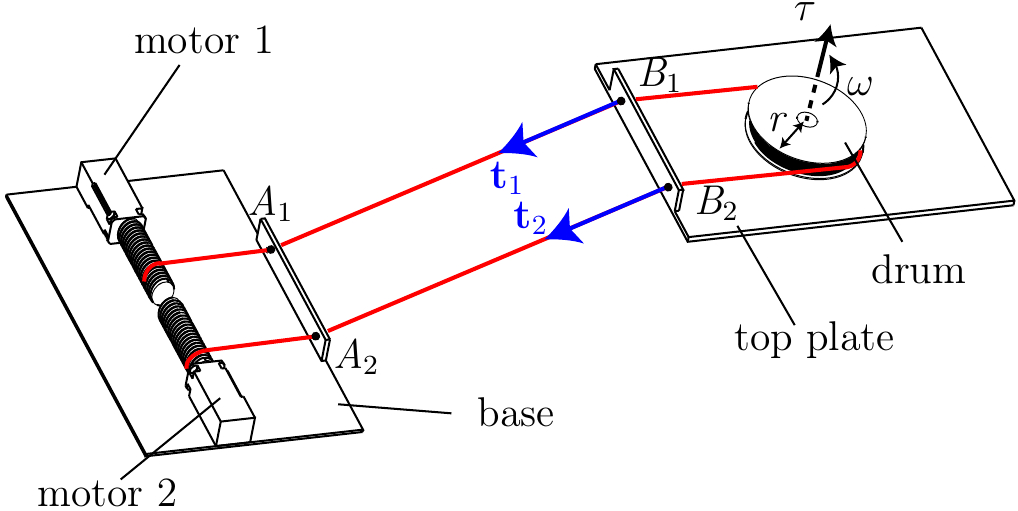}
			\caption{Representation of a cable-loop}
			\label{fig:cable_loop}
		\end{center}
	\end{figure}

	Figure~\ref{fig:cable_loop} represents a simplified form of the cable-loop system. The latter consists of a single cable of which both ends are actuated by two motors while passing through exit-points, namely, $\textit{A}_1$, $\textit{A}_2$, and anchor-points, namely, $\textit{B}_1$, $\textit{B}_2$. The cable-loop is coiled up around a drum on the moving platform. The cable-loop drum then acquires one rotational DOF with respect to the moving platform. These can be used to actuate an embedded mechanism or to control additional degrees of freedom such as rotations over wide ranges. The purpose of the cable-loop is double. Firstly, its aim is to translate the top plate as two single cables would do when the coiling directions of both motors are the same. Secondly, it actuates the embedded drum by circulating the cable when coiling directions of the two actuators are different.\par
	
	By controlling the difference of tension in both ends of the cable loop, namely $\delta t_{12} = t_1 - t_2$, it is possible to transmit torque $\tau$ to the pulley. This capacity can be used to increase the orientation workspace of the end-effector when the pulley is used to rotate the end-effector or the pulley can actuate an embedded mechanism without having the drawbacks of embedding the actuators on the platform. Here, three cable loops are used to actuate independently the three omni-wheels in contact with the sphere as illustrated in Fig.~\ref{fig:cable_loop}.\par
	
	\subsection{Overall Parameterization}\label{subsec:DESCRIPTION_overall_parametrization}
	
	As shown in Fig.~\ref{fig:PSW2_CREATOR_detail}, $\mathcal{F}_0$ denotes the frame fixed to the base of origin point $\textit{O}_0$ and axes $\vec{x}_0$, $\vec{y}_0$ and $\vec{z}_0$. $\mathcal{F}_1$ is the frame attached to the top-plate of origin point $\textit{O}_1$ and axes $\vec{x}_1$, $\vec{y}_1$ and $\vec{z}_1$. $\mathcal{F}_2$ is the frame attached to the end-effector, i.e., the sphere, of origin point $\textit{O}_2$, the geometric center of the sphere and axes $\vec{x}_2$, $\vec{y}_2$ and $\vec{z}_2$.\par
	
	It is noteworthy that $\mathcal{F}_0$ and $\mathcal{F}_1$ have a translational and orientational relative movement while $\mathcal{F}_1$ and $\mathcal{F}_2$ only have a relative rotational movement.\par
	
	The exit points $A_i$ are the points belonging to the frame through which the cables are routed between the winch and the top plate. The anchor points $B_i$ are the points belonging to the TP where the cables are connected. It is noteworthy that the cables are connecting exit points and anchor points accordingly and a unit vector expresses the cable direction. Therefore, the loop-closure equations associated with each cable are expressed as follows:
	
	\begin{equation}\label{eq:cab_length}
		\begin{aligned}
			{^0}\mathbf{l}_{i} = {^0}\mathbf{a}_{i} - {^0}\mathbf{p} - {^0}\mathbf{R}_{1}{^1}\mathbf{b}_{i}
		\end{aligned}
	\end{equation}
	
	\noindent with $ i \in [\![1,\dots,8]\!]$ where ${^0}\mathbf{l}_{i}$ is the $i$-th cable vector, ${^0}\mathbf{a}_{i}$ is the corresponding anchor point expressed in the base frame, ${^1}\mathbf{b}_{i}$ is the coordinate vector of exit point in the platform frame,  ${^0}\mathbf{p}$ is the position vector of the platform frame and ${^0}\mathbf{R}_{1}$ is the rotation matrix from $\mathcal{F}_0$ to $\mathcal{F}_1$.
	
	We can then write the $i$-th unit cable vector as:
	
	\begin{equation}
		{^0}\mathbf{u}_{i} = \dfrac{{^0}\mathbf{l}_{i}}{l_{i}}
	\end{equation}
	
	\noindent with $l_{i}$ being the $i$-th cable length.
	
	\section{Kinetostatic Model of the Manipulator}\label{sec:KINETOSTATIC}
	
	In this section, we proceed to the kinetostatic modelling of the overall manipulator. We write the static equation of the manipulator as follows:
	
	\begin{equation}
		\mathbf{W} \mathbf{t} + \mathbf{w}_{e} = \mathbf{0}_{9}
	\end{equation}	
	
	\noindent with $\mathbf{W}$ being the wrench matrix, $\mathbf{t}$ being the cable tension vector and $\mathbf{w}_{e}$ being the external wrenches applied on the platform. In our case, we only consider the action of the weight of the moving platform as external wrench.\par
	
	The wrench matrix of the manipulator $\mathbf{W}$ is the concatenation of the wrench matrices of both mechanisms:  
	
	\begin{equation}\label{eq:W}
		\mathbf{W} = 
		\begin{bmatrix}
			\mathbf{W}_{TP}\\
			\mathbf{W}_{SW}
		\end{bmatrix}_{9\times 8}
	\end{equation}
	
	\noindent where $\mathbf{W}_{TP}$ is the wrench matrix associated to the top plate and $\mathbf{W}_{SW}$ is the wrench matrix related to the PSW.\par
	
	Similarly, the wrench vector of the manipulator consists of the wrench vector of both mechanisms:
	
	\begin{equation}\label{eq:w_g}
		\mathbf{w}_g = 
		\begin{bmatrix}
			\mathbf{w}_g^{TP}\\
			\mathbf{w}_g^{SW}
		\end{bmatrix}_{9\times 1}
	\end{equation}
	
	\noindent where $\mathbf{w}_g^{TP}$ is the wrench vector associated to the top plate and $\mathbf{w}_g^{SW}$ is the wrench vector exerted on the PSW. $\mathbf{W}_{TP}$ and $\mathbf{w}_g^{TP}$ are defined in Section~\ref{subsec:KINETOSTATIC_TP} while $\mathbf{W}_{SW}$ and $\mathbf{w}_g^{SW}$ are defined in Section~\ref{subsec:KINETOSTATIC_SW}.\par
	
	\subsection{Kinetostatic Model of the Top Plate}\label{subsec:KINETOSTATIC_TP}
	
	In this section we write the kinetostatic model of the top plate. The static equilibrium of the platform can be written as:
	
	\begin{equation}
		\mathbf{W}_{TP} \mathbf{t} + \mathbf{w}_g^{TP} = \mathbf{0}_{6}
	\end{equation}
	
	$\mathbf{W}_{TP}$ takes the following form:
	
	\begin{equation}\label{eq:W_TP}
		\mathbf{W}_{TP} = 
		\begin{bmatrix}
			{^0}\mathbf{u}_{1} & {^0}\mathbf{u}_{2} & {^0}\mathbf{u}_{3} & {^0}\mathbf{u}_{4} & {^0}\mathbf{u}_{5} & {^0}\mathbf{u}_{6} & {^0}\mathbf{u}_{7} & {^0}\mathbf{u}_{8}\\
			{^0}\mathbf{d}_{1} & {^0}\mathbf{d}_{2} & {^0}\mathbf{d}_{3} & {^0}\mathbf{d}_{4} & {^0}\mathbf{d}_{5} & {^0}\mathbf{d}_{6} & {^0}\mathbf{d}_{7} & {^0}\mathbf{d}_{8}\\
		\end{bmatrix}_{6\times 8}
	\end{equation}
	
	\noindent with ${^0}\mathbf{d}_i$ being the cross-product of vectors $\mathbf{b}_i$ and $\mathbf{u}_i$ expressed in the base frame $\mathcal{F}_0$ as:
	
	\begin{equation}
		{^0}\mathbf{d}_{i} =  {^0{\mathbf{R}}_1}{^1\mathbf{b}_i} \times {^0}\mathbf{u}_i
	\end{equation}
	
	\noindent We define the wrench vector of the top plate $\mathbf{w}_{TP}$ as follows:
	
	\begin{equation}
		\mathbf{w}_g^{TP} =
		\begin{bmatrix}
			\mathbf{f}_{TP}\\
			\mathbf{m}_{TP}\\
		\end{bmatrix}_{6\times 1}
	\end{equation}
	
	\noindent where $\mathbf{f}_{TP}$ is the 3-dimensional force and $\mathbf{m}_{TP}$ is the 3-dimensional moment exerted on the top plate.
	
	\subsection{Kinetostatic Model of the Parallel Spherical Wrist}\label{subsec:KINETOSTATIC_SW}
\begin{figure}[!htbp]
		\centering
		\includegraphics[width=1\linewidth]{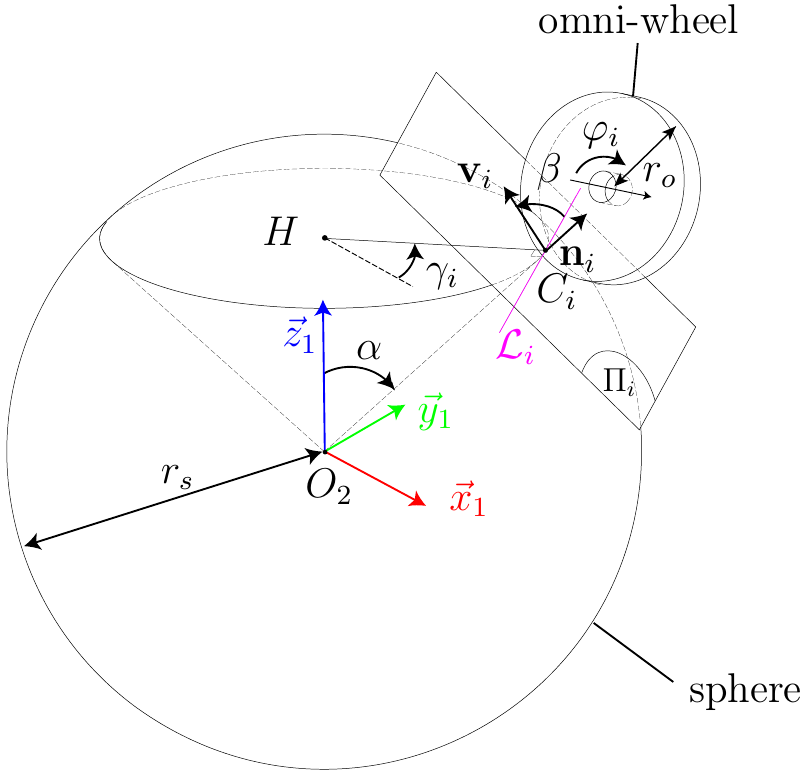}
		\caption{Parametrization of the Parallel Spherical Wrist}
		\label{fig:PSW_parametrization}
	\end{figure}

	The parametrization of the PSW is illustrated in Fig.~\ref{fig:PSW_parametrization} and described below:
	\begin{description}
		\item $\textit{C}_i$: Contact point between the $i$th omni-wheel and the sphere
		\item $\Pi_i$: Plane passing through the contact point~$C_i$ and tangent to the sphere
		\item $\alpha$: Elevation angle of point $C_i$, $ \alpha \in [0,\pi] $ 
		\item $\beta$: Angle between the tangent line~$\mathcal{L}_i$ and the actuation force of the omni-wheel, $\beta \in [-\frac{\pi}{2},\frac{\pi}{2}]$
		\item $\gamma_i$: Angle between $\vec{x}_1$ and the vector pointing from point~$H$ to point~$C_i$, $i=1,2,3$ 
		\item $\textit{r}_s$: Sphere radius
		\item $\textit{r}_o$: Omni-wheel radius
		\item $\dot{\varphi}_i$: Angular velocity of the $i$-th omni-wheel
		\item $\mathbf{v}_i$: Unit vector along the actuation force produced by the $i$-th omni-wheel on the sphere
		\item $\mathbf{n}_i$: Unit vector normal to plane~$\Pi_i$
	\end{description}
	
	The optimal set of the parameters of the wrist was defined in~\cite{Platis.17} in order to maximize the amplitudes of its orientation as well as its dexterity. The following hypotheses are taken into account for the wrist modelling and analysis: ($i$)~The omni-wheels are normal to the sphere; ($ii$)~The contact points between the omni-wheels and the sphere belong to the circumference of a circle. The latter is the base of an inverted cone, its tip being the centre of the sphere. The angle between the vertical axis and the cone is named~$\alpha$; ($iii$)~In the plane containing the cone base, the three contact points form an equilateral triangle.
	
	The angular velocity vector of the sphere $\omega= [\omega_x, \omega_y, \omega_z]^\mathsf{T} $ is expressed as a function of the angular velocity vector of the omni-wheels $ \dot{\varphi}= [\dot{\varphi_1}, \dot{\varphi_2}, \dot{\varphi_3}]^\mathsf{T}$, as follows:
	\begin{equation}
		\mathbf{A}\omega = \mathbf{B} \dot{\varphi}
		\label{eq:AtBPhi}
	\end{equation}
	$\mathbf{A}$ and $\mathbf{B}$ are the forward and inverse Jacobian matrices of the PSW, which take the form:
	\begin{equation}\label{eq:Avec}
		\mathbf{A} = r_s\begin{bmatrix}
			(\mathbf{n}_1 \times \mathbf{v}_1)^\mathsf{T}\\
			(\mathbf{n}_2 \times \mathbf{v}_2)^\mathsf{T}\\
			(\mathbf{n}_3 \times \mathbf{v}_3)^\mathsf{T}\\
		\end{bmatrix}
	\end{equation}
	and,
	\begin{equation}
	\mathbf{B} = \textit{r}_o \mathbf{1}_{3\times3}
	\end{equation}	
	
	From Eq.~(\ref{eq:Avec}), matrix~$\mathbf{A}$ is expressed as a function of angles $\alpha$, $\beta$ and $\gamma_i$ as:
	\begin{equation}
		\mathbf{A} = r_s\begin{bmatrix}
			S_\beta C\gamma_1 - C\alpha C\beta C\gamma_1 & -S_\beta C\gamma_1 - C\alpha C\beta C\gamma_1 & S_\alpha C_\beta\\
			S_\beta C\gamma_2 - C\alpha C\beta C\gamma_2 & -S_\beta C\gamma_2 - C\alpha C\beta C\gamma_2 & S_\alpha C_\beta\\
			S_\beta C\gamma_3 - C\alpha C\beta C\gamma_3 & -S_\beta C\gamma_3 - C\alpha C\beta C\gamma_3 & S_\alpha C_\beta
		\end{bmatrix}
	\end{equation}
	
	\noindent where $C_{k}=\cos(k)$ and $S_{k}=\sin(k)$, $k=\alpha, \beta, \gamma_1, \gamma_2, \gamma_3$. From~\cite{Platis.17}, the PSW is fully isotropic from a kinematic viewpoint if and only if $\alpha = 35.2^\circ$ and $\beta = 0^\circ$. Therefore, those values are selected in what remains. Equation~(\ref{eq:AtBPhi}) is rewritten as follows:
			
	\begin{equation}
		\omega = \mathbf{J}_\omega \dot{\varphi}
		\label{eq:OmegaJPhi}
	\end{equation}
	
	\noindent where $\mathbf{J}_\omega = \mathbf{A}^{-1}\mathbf{B}$, is the Jacobian matrix of the wrist, i.e., $ \mathbf{J}_\omega$ is the mapping from angular velocities of the omni-wheels into the required angular velocity of the end-effector. 
	Based on the theory of reciprocal screws~\cite{Ball.00}, it turns out that:
	
	\begin{equation}
		\mathbf{m}_{SW}^\mathsf{T}\omega = \tau^\mathsf{T} \dot{\varphi}
		\label{eq:mOmegaTauPhi}
	\end{equation}
	
	\noindent where $\mathbf{m}_{SW} = [m_x, m_y, m_z]^\mathsf{T}$ is the output moment vector of the sphere and $\tau = [\tau_1, \tau_2, \tau_3]^\mathsf{T}$ is the input torque vector, namely the omni-wheel torque vector. By substituting Eq.~(\ref{eq:OmegaJPhi}) into Eq.~(\ref{eq:mOmegaTauPhi}), we obtain:
	
	\begin{equation}
		\tau = \mathbf{J}_\omega^\mathsf{T} \mathbf{m}_{SW} = \mathbf{W}_{\omega} \mathbf{m}_{SW}
	\end{equation}\label{eq:TauJm}
	
\begin{figure}[!htbp]
		\centering
		\includegraphics[width=1\linewidth]{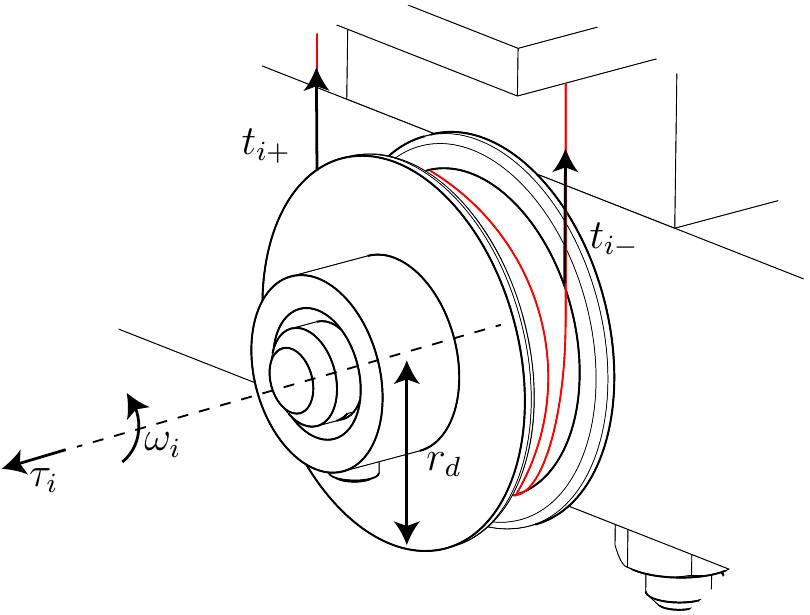}
		\caption[Torque]{Representation of a cable-loop drum}
		\label{fig:drum_cl_detail}
	\end{figure}

	The wrench matrix $\mathbf{W}_{\omega} = \mathbf{J}_\omega^\mathsf{T}$ maps the output torque of the sphere to the omni-wheels torques. As shown in Fig.~\ref{fig:drum_cl_detail}, $\tau$ can be expressed as a function of the cable tensions such that:
	\begin{eqnarray}\label{eq:TauiRiTij}
		\tau_1 = r_d(t_1 - t_2)\label{eq:TauiRiTij1}\\ 
		\tau_2 = r_d(t_3 - t_4)\label{eq:TauiRiTij2}\\ 
		\tau_3 = r_d(t_5 - t_6)\label{eq:TauiRiTij3}
	\end{eqnarray} where $r_d$ is the radius of the embedded drum of the cable-loops.
	
	From Eqs.~(\ref{eq:TauiRiTij1}) to~(\ref{eq:TauiRiTij3}), the omni-wheel torque vector~$\tau$ is expressed as a function of the cable tensions as follows:
	\begin{equation}
		\tau = \mathbf{W}_{c} \mathbf{t}
	\end{equation}\label{eq:TauWt}
	where $\mathbf{t} = [t_1,\, \dots,\, t_8]^\mathsf{T}$ is the cable tension vector and $\mathbf{W}_{c}$ is the matrix assigning the cables to the cable-loops. If cable pairs ($1$, $2$), ($3$, $4$) and ($5$, $6$) respectively correspond to cable-loops~$1$,é $2$ and~$3$, $\mathbf{W}_{c}$ will take the form:
	\begin{equation}
		\mathbf{W}_{c} =
		\begin{bmatrix}
			r_d & -r_d & 0 & 0 & 0 & 0 & 0 & 0\\
			0 & 0 & r_d & -r_d & 0 & 0 & 0 & 0\\
			0 & 0 & 0 & 0 & r_d & -r_d & 0 & 0 
		\end{bmatrix} 
	\end{equation}
	
	The equilibrium of the wrench applied on the PSW is expressed as:
	\begin{equation}
		\mathbf{m}_{SW} = \mathbf{W}_{SW} \mathbf{t}
	\end{equation}
	with $\mathbf{m}_{SW}$ being the external moments applied by the environment onto the PSW. The wrench matrix~$\mathbf{W}_{SW}$ expresses the relationship between the cable tensions and the moment applied on the wrist.
	\begin{equation}\label{eq:W_SW}
		\mathbf{W}_{SW} = \mathbf{W}_{\omega} \mathbf{W}_{c}
	\end{equation}

	The orientation of the sphere with respect to the base frame is defined by the pitch angle~$\theta$, the yaw angle~$\psi$ and the roll angle~$\chi$ while following the ZYX-Euler-angles convention. Those three angles are the components of the orientation vector~$\mathbf{q}_{SW}$ of the wrist, namely,
	\begin{equation}
		{\mathbf{q}}_{SW}=
		\begin{bmatrix}
			\theta\\ 
			\psi\\ 
			\chi
		\end{bmatrix}
	\end{equation}
	
	\section{Workspace Analysis}\label{sec:WORKSPACE}
	
	The static workspace of the manipulator consists of the set of positions and orientations of the moving platform and the orientations of the end-effector, namely, ${^0\mathbf{p}}$ and ${^0\mathbf{R}_{1}}$ and $^0\mathbf{q}_{SW}$, which satisfies the static equilibrium of the manipulator.\par
	
	The cable tension set $\mathcal{T}$ amounts to a hyper-cube in an eight-dimensional space:
	
	\begin{equation}
		\mathcal{T} = \{\mathbf{t} \in \mathbb{R}^{8} : \mathbf{t}_{min} \leq \mathbf{t} \leq \mathbf{t}_{max}\}
	\end{equation}
	
	\noindent  where $\mathbf{t}_{min}$ and $\mathbf{t}_{max}$ are respectively the lower and upper bounds of the cable tension.
	
	The static workspace of the manipulator is defined as follows:
	
	\begin{multline}
		\mathcal{S} = \{{(^0\mathbf{p}},{^0\mathbf{R}_{1}}, {^0}\mathbf{q}_{SW} \in\mathbb{R}^{3}\times SO(3) \times \mathbb{R}^{3}: \exists\mathbf{t}\in\mathcal{T}, \\ \mathbf{W}\mathbf{t} + \mathbf{w}_{e} = \mathbf{0}_{9}\}
		\label{eq:SW_3T3R}
	\end{multline}
	
	\noindent where $SO(3)$ is the group of proper rotation matrices. The static workspace in a nine-dimensional space is expressed using the Equation~(\ref{eq:SW_3T3R}). As the visualization of such high-dimensional space is impossible with common human perception in 3D, we define the static workspace of the manipulator for a simplified case. We define $\mathcal{S}$ for a constant orientation of the top-plate. The former subset, namely, $\mathcal{S}_{AO}$ is a set for a given orientation of the top-plate while the wrist is free to rotate:\par
	
	\begin{multline}
		\mathcal{S}_{AO} = \{ {^0\mathbf{p}} \in\mathbb{R}^{3}\,|\,{^0{\mathbf{R}}_1}= \mathbf{I}_3|  -\pi\leq \theta, \;\psi,\; \chi \leq \pi :\exists\mathbf{t}\in\mathcal{T}, \\ \mathbf{W}\mathbf{t} + \mathbf{w}_{e} = \mathbf{0}_{9}\}
		\label{SW_AO}
	\end{multline}
	
	The discretization of the Cartesian space is made so that $n_x$, $n_y$ and $n_z$ are the numbers of discretized points along  $\vec{x}_0$, $\vec{y}_0$ and $\vec{z}_0$ axes, respectively.
	
	$\mathcal{R}_{\mathcal{S}}$ is defined as the proportion of the static workspace to the overall space occupied by the manipulator:
	
	\begin{equation}
		\mathcal{R}_{\mathcal{S}} = \frac{N_{\mathcal{S}}}{(n_x +1)(n_y + 1)(n_z + 1)}
	\end{equation}
	
	\noindent with $N_{\mathcal{S}}$ being the number of points inside the discretized static workspace $\mathcal{S}$.\par
	
	\section{Optimal Cable Arrangement}\label{sec:ARRANGEMENT}
	
	This section deals with the cable arrangement of the manipulator for obtaining the maximum size of the static workspace. The $i$-th cable arrangement is the association of anchor points $B_i$ to exit points $A_i$. The top plate has fifteen points for anchor points and the manipulator has a maximum of eight actuators. Therefore, the optimal cable arrangement is defined as the association of eight anchor points to eight exit points such that the workspace is maximized.\par
	
	\begin{figure}[!htbp]
		\centering
		\includegraphics[width=1\linewidth]{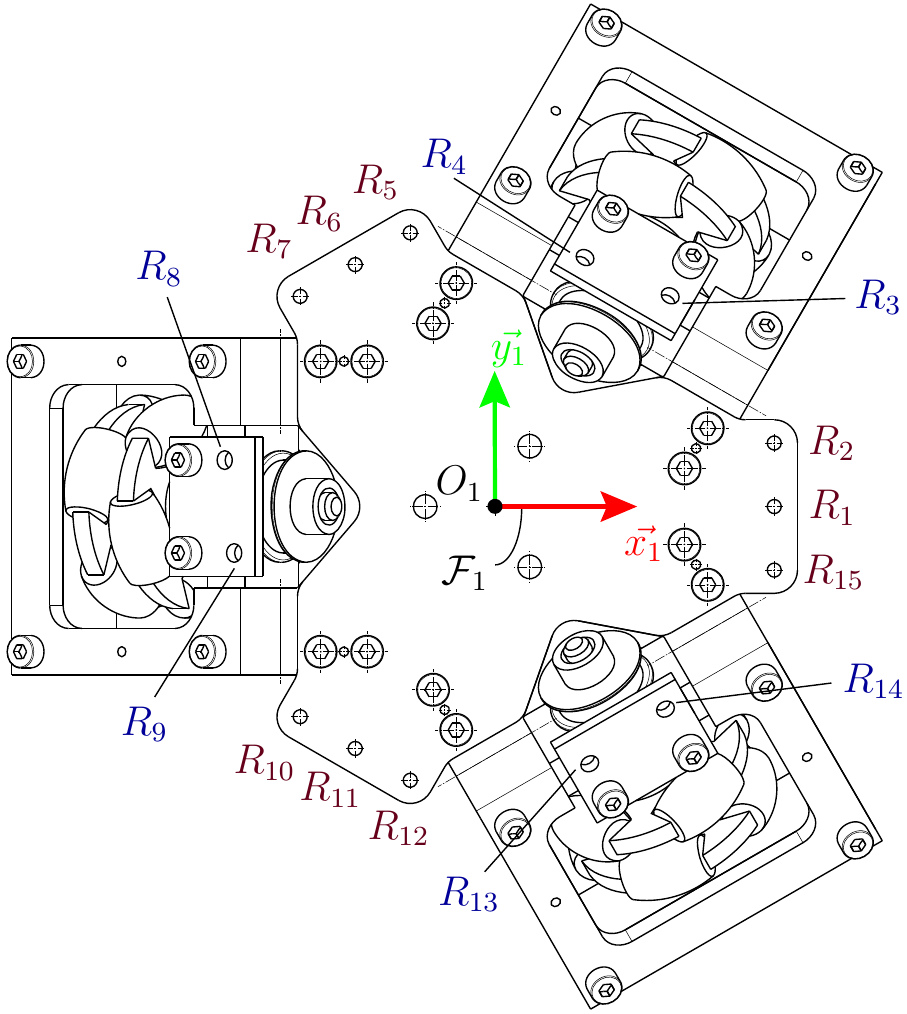}
		\caption[Anchor]{The fifteen points for the anchor points on the moving-platform}
		\label{fig:anchor_points}
	\end{figure}
	
	The number of exit point combinations, $\mathcal{N}_e$ is:
	
	\begin{equation}
		\mathcal{N}_e = \dbinom{n_e}{n_c}
	\end{equation}\label{eq:n_e}
	
	\noindent with $\textit{n}_e$ and $\textit{n}_c$ being the numbers of available exit-points and cables, respectively. The number of anchor-points combinations, $\mathcal{N}_a$, consists in the number of permutations of the set of points, which is given by:
	
	\begin{equation}
		\mathcal{N}_a = \dbinom{n_a}{n_c} n_c!
	\end{equation}\label{eq:n_a}
	
	\noindent $\textit{n}_a$ being the number of selected anchor points. $\mathcal{S}_C$ is the set of possible cable configuration, the number of cable configuration $\mathcal{N}_C = dim(\mathcal{S}_C)$ is thus given by:
	
	\begin{equation}
		\mathcal{N}_c = \mathcal{N}_a \mathcal{N}_e = \dbinom{n_e}{n_c} \dbinom{n_a}{n_c} n_c!
	\end{equation}\label{eq:N_c}
	
	Figure~\ref{fig:anchor_points} shows all the available anchor points, namely, $\textit{r}_i$, $i=1,~2,~\dots,~15$, which are divided into two groups: $\mathcal{S}_{CL}  = \{R_{3}, R_{4}, R_{8}, R_{9}, R_{13}, R_{14}\}$ which is the set of points associated to cable-loops and  $\mathcal{S}_{SC}  = \{R_{1}, R_{2}, R_{5}, R_{6}, R_{7}, R_{10}, R_{11}, R_{12}, R_{15}\}$ which is the set of points associated to the simple cables.
	Six points amongst the fifteen points are selected to make the three cable loops. Therefore, nine remaining anchor points host the remaining two single-actuated cables.\par
	
	\begin{equation}
		n_c = n_{SC} + n_{CL}
	\end{equation} 
	
	The number of available anchor points for the single and bi-actuated cables are denoted as $n_{aSC}$ and $n_{aCL}$:
	
	\begin{equation}
		n_a = n_{aSC} + n_{aCL}
	\end{equation}
	
	The number, $\mathcal{N}_{CL}$, of combinations considering cable-loop is given by:
	
	\begin{equation}
		\mathcal{N}_{CL} = \dbinom{n_e}{n_c} \dbinom{n_{aSC}}{n_{SC}} n_c!
		\label{eq:N_CL}
	\end{equation}
	
	Six points are associated to the wrist actuation and consequently to the three cable-loops: $n_{aCL} = 6$. The remaining nine anchor points are to be assigned to simple cables $n_{aSC} = 9$. Finally, $\mathcal{N}_{CL}$, expressed in Eq.~(\ref{eq:N_CL}) is given by:
	
	\begin{equation}
		\mathcal{N}_{CL} = \binom{8}{8} \binom{9}{2} 8! = 1~451~520
	\end{equation}
	
	As $\mathcal{N}_{CL}$ is very large, for computing-time sake, the number of available points for anchor points is supposed to be equal to three. Thus $\mathcal{S}_{SC} = \{R_{1},R_{6},R_{11}\}$. By substituting $n_{aSC} = 3$ into Eq.~(\ref{eq:N_CL}) we obtain:
	
	\begin{equation}
		\mathcal{N}_{CL} = \binom{8}{8} \binom{3}{2} 8! = 120~960
	\end{equation}
	
\begin{figure}[!htbp]
		\centering
		\subfloat[1][Cable configuration]{
			\includegraphics[width=1\linewidth]{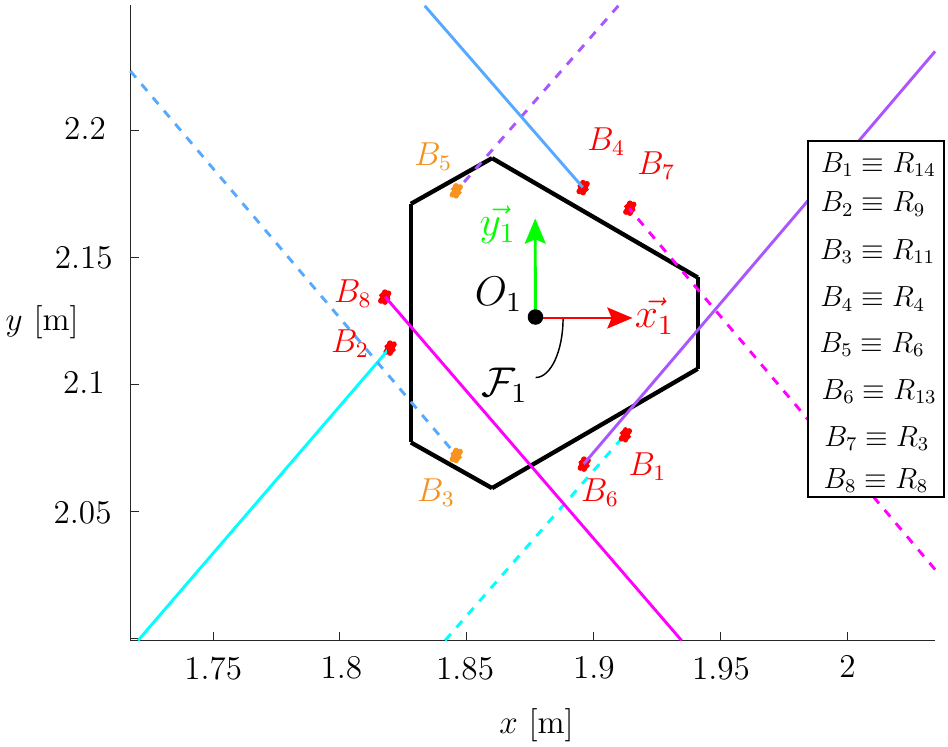}
			\label{subfig:cable_arrangement}}
		\qquad
		\subfloat[2][Static workspace]{
			\includegraphics[width=1\linewidth]{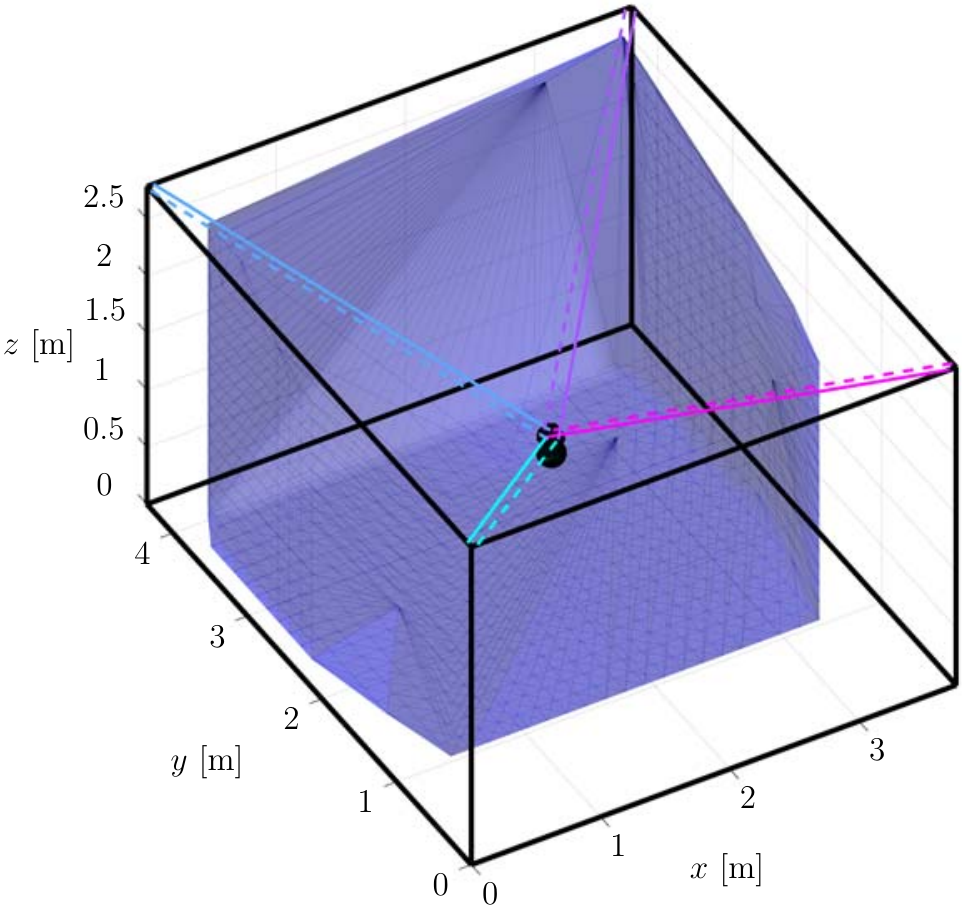}
			\label{subfig:workspace}}	
		\caption{Optimum cable arrangement and static workspace}
		\label{fig:simulation}
	\end{figure}

	The static workspace is computed for the 120~960 cable arrangements. Figure~\ref{fig:simulation} illustrates the cable arrangement corresponding to the largest workspace. Figure~\ref{subfig:cable_arrangement} shows a schematic of the anchor-plate as well as the optimal cable arrangement. Figure~\ref{subfig:workspace} shows the corresponding static workspace with $\mathcal{R}_{\mathcal{S}} = 65 \%$.\par
	
	\section{Prototyping and Experimentation}\label{sec:DEMONSTRATOR}
\begin{figure}[!htbp]
		\centering
		\includegraphics[width=\linewidth]{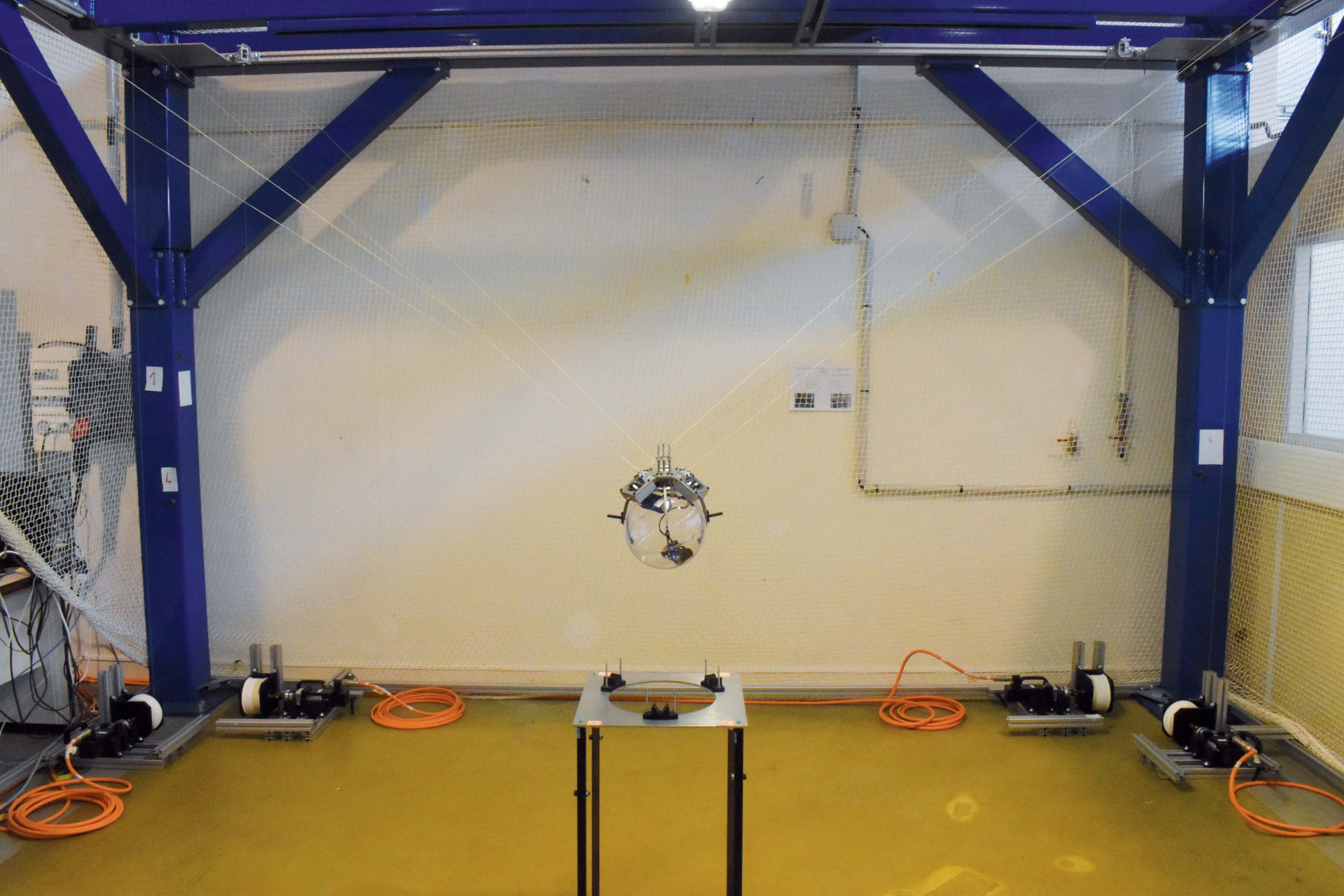}
		\caption[control]{The CDPR prototype with full-circle end-effector rotations}
		\label{fig:CDPR_FCEE_proto}
	\end{figure}

\begin{figure}[!htbp]
		\centering
		\includegraphics[width=\linewidth]{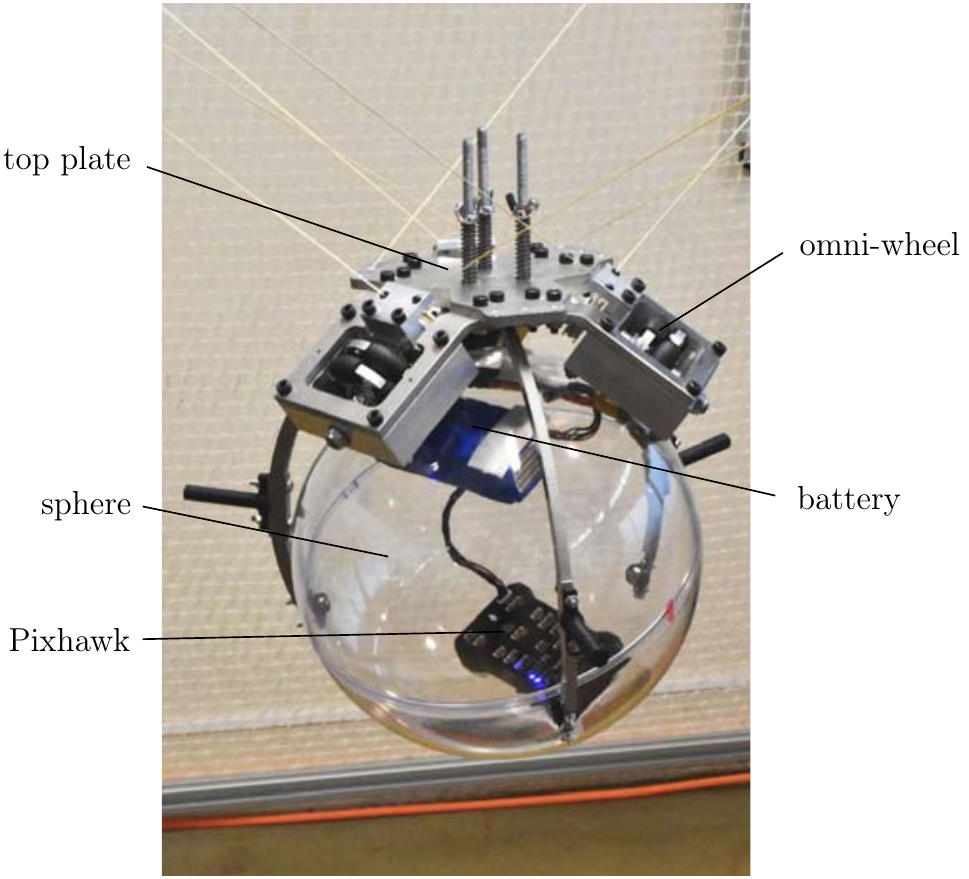}
		\caption[Prototype]{The parallel spherical wrist equipped with a Pixhawk flight controller}
		\label{fig:exp_proto}
	\end{figure}

	The prototyping of a CDPR with full-circle end-effector rotations is presented in this section. The base frame of the prototype shown in Fig.~\ref{fig:CDPR_FCEE_proto} is 4~m long, 3.5~m wide and 4~m high. The full-circle end-effector rotations are obtained thanks to the PSW shown in Fig.~\ref{fig:exp_proto}. The PSW is mainly made up of a top plate, three double omni-wheels and a transparent sphere. The three-DOF rotational motions of the sphere are obtained from the rotations of the omni-wheels. Each omni-wheel is driven by a cable loop, the strands of the cable being respectively wound around two actuated reels fixed to the ground. A flight controller \textsuperscript{\textcopyright}Pixhawk, embedded in the sphere, is used to measure its orientation, angular velocity and linear acceleration. This controller, equipped with a gyroscope, an accelerometer and a magnetometer, acts as a data-logger device to record measurements. The overall mass of the PSW is equal to 1.87~kg.  It should be noted that the manipulator is under-actuated because the moving-platform including the PSW has nine degrees of freedom whereas the prototype has only eight actuators.
	
    \begin{figure}[!b]
   		\centering
   			\includegraphics[width=\linewidth]{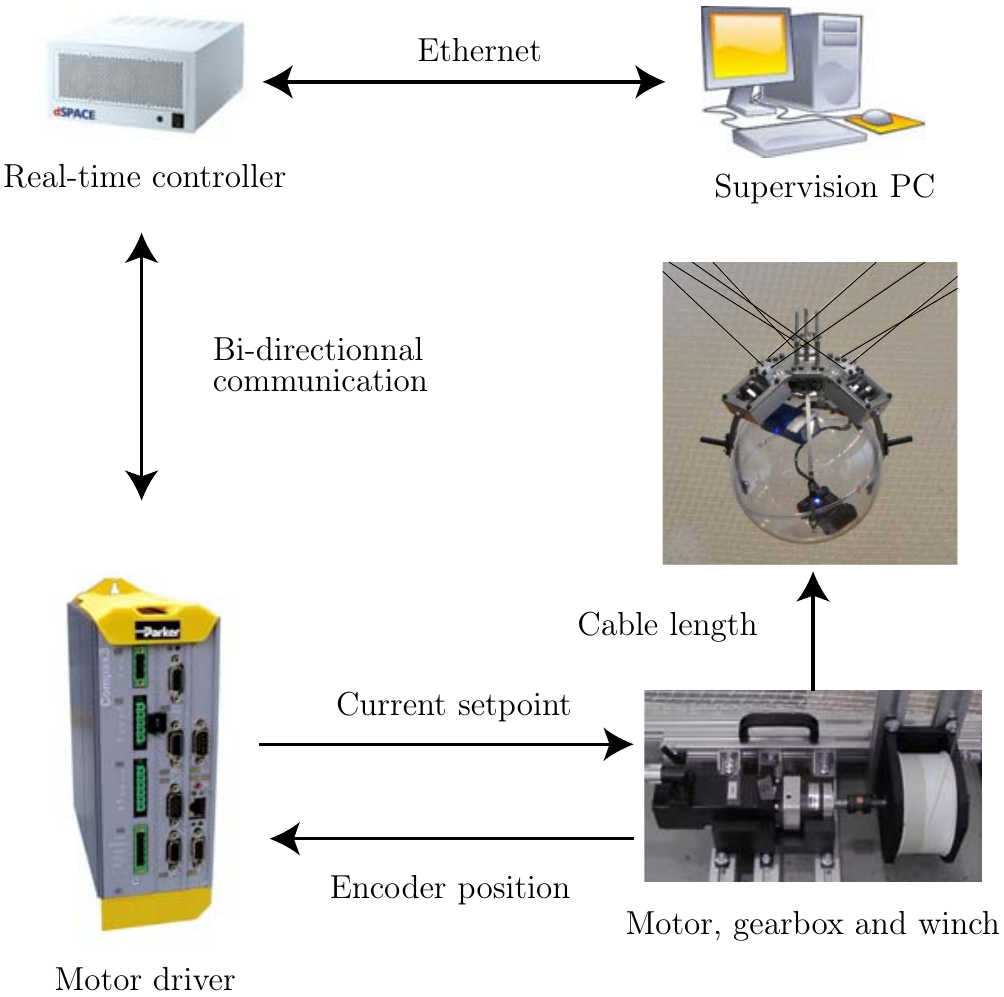}
   		\caption[control]{Equivalent architecture of the prototype}
   		\label{fig:CREATOR_control}
	\end{figure}

Figure~\ref{fig:CREATOR_control} shows the main hardware of the prototype, which consists of a PC (equipped with \textsuperscript{\textcopyright}MATLAB and \textsuperscript{\textcopyright}ControlDesk software), eight \textsuperscript{\textcopyright}PARKER SME60 motors and TPD-M drivers, a \textsuperscript{\textcopyright}dSPACE DS1007-based real-time controller and eight custom-made winches.

Slippage between the omni-wheels and the sphere inevitably leads to drift in the orientation when tracking a prescribed trajectory. Therefore, this robot is intended for teleoperation applications where an operator can compensate for the errors that accumulate over time. In this context, the robot performance is best assessed by comparing its angular velocity response to an angular velocity input rather than by looking at its ability to track position and orientation over a trajectory.\\

The performances of the protoype were experimentally evaluated along the following three trajectories:
\begin{description}
	\item[Trajectory~1:] Pure rotational motions of the sphere about axes parallel to $x_0$, $y_0$ and $z_0$, respectively, while the top plate is fixed to its support. A fifth-order polynomial is used to obtain continuous velocity and acceleration trajectory profiles. 
	\item[Trajectory~2:] Pure translational motions of the top plate along four successive straight line segments whereas the sphere does not rotate. A fifth-order polynomial is used to determine the velocity and acceleration profiles along each line segment.
	\item[Trajectory~3:] The top plate performs a vertical translational motion while the sphere rotates.
\end{description}   

\begin{figure}[!htbp]
	\centering
	\subfloat[about $x_0$~axis]{
		\centering
		\includegraphics[clip,width=1\linewidth]{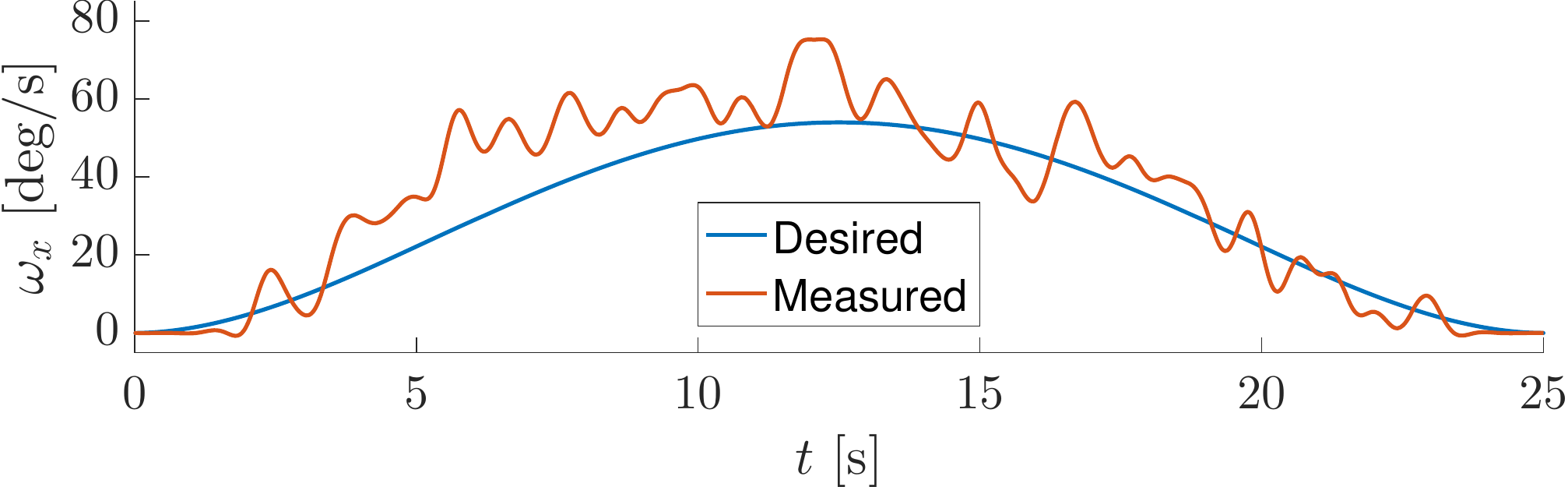}
		\label{fig:omega_x_analysis}
	}
	
	\subfloat[about $y_0$~axis]{%
		\centering
		\includegraphics[clip,width=1\linewidth]{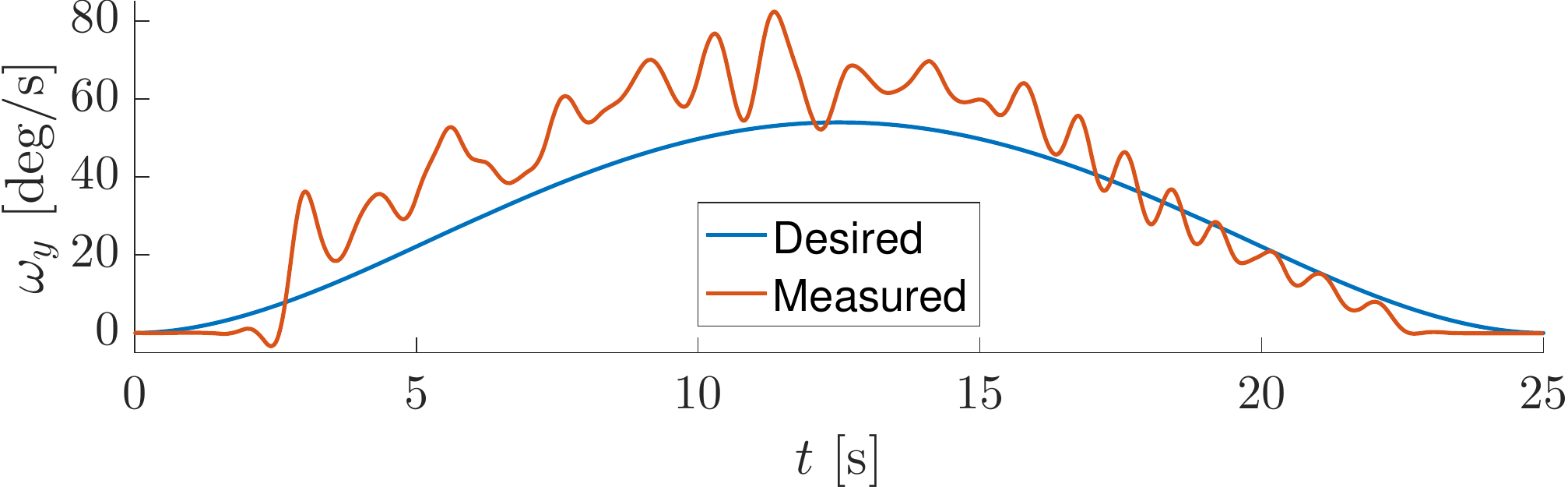}
		\label{fig:omega_y_analysis}
	}
	
	\subfloat[about $z_0$~axis]{
		\centering
		\includegraphics[clip,width=1\linewidth]{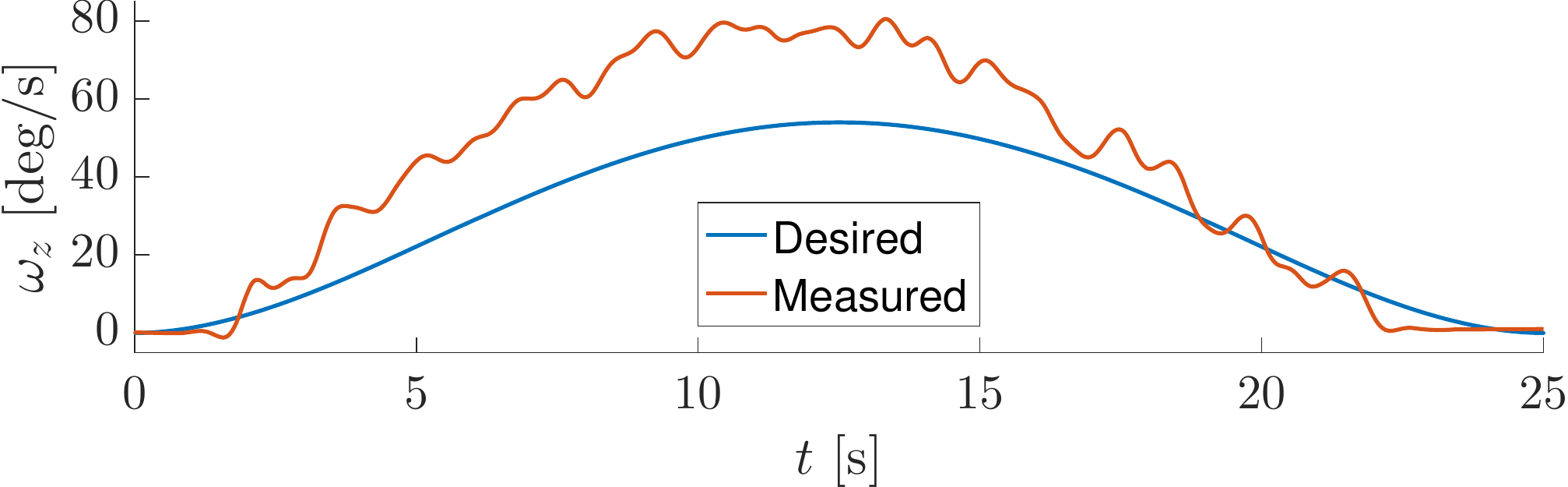}
		\label{fig:omega_z_analysis}
	}
	\caption{Desired and measured angular velocities of the sphere along Trajectory~1}
	\label{fig:omega_analysis}
\end{figure}

A \href{https://bit.ly/3hvpsL2}{video}\footnote[1]{\url{https://bit.ly/3hvpsL2}} illustrates the experiments carried out with the prototype. The experimental motions of the sphere along Trajectory~1 are shown in the video from 0min~5s to 1min~7s. Figures~\ref{fig:omega_analysis}(a)-(c) depict the difference between the desired and measured angular velocities~$\omega_x$ ($\omega_y$, $\omega_z$, resp.) about axis~$x_0$ ($y_0$, $z_0$, resp.) of the sphere expressed in frame~$\mathcal{F}_0$.

\begin{figure}[!htbp]
	\centering
	\includegraphics[width=1\linewidth]{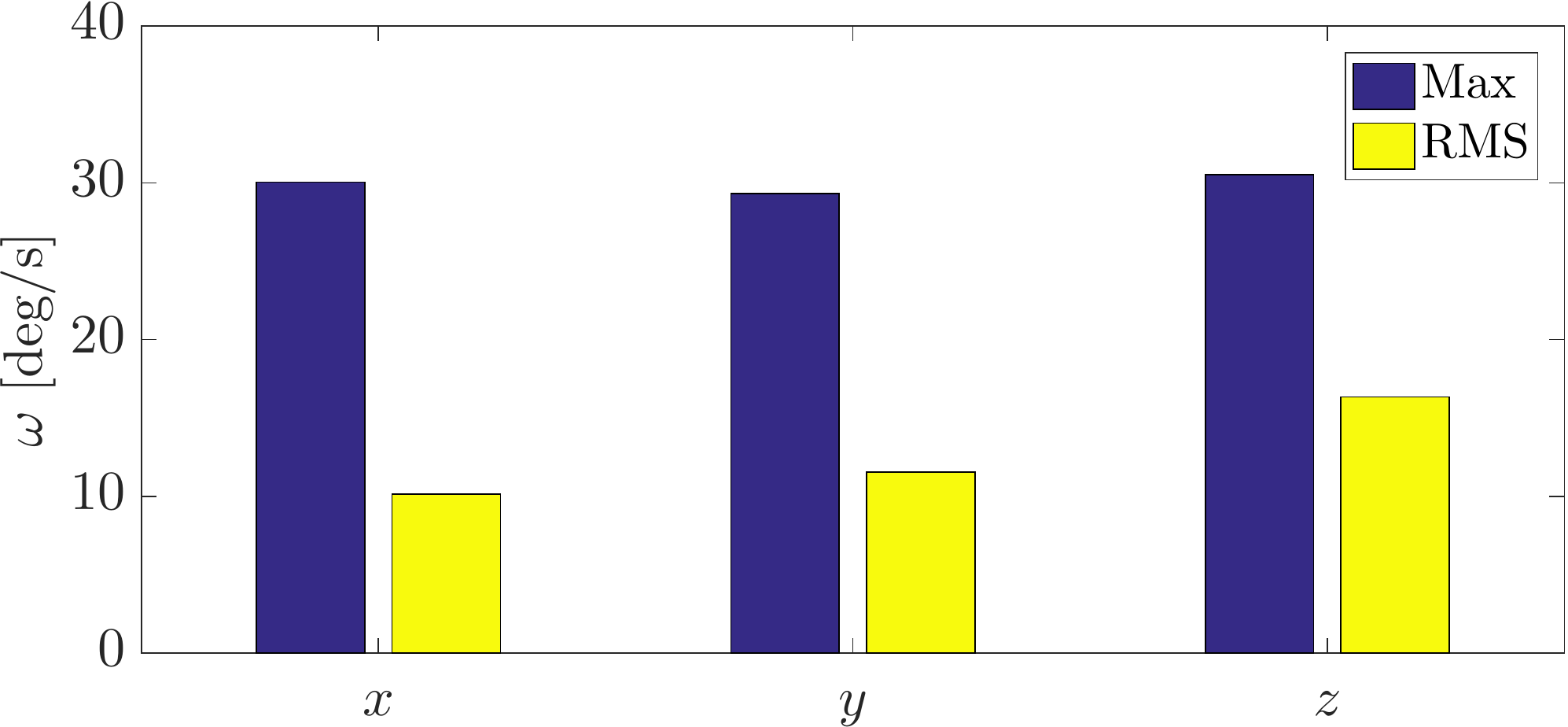}
	\caption{Maximum and root mean square errors on the sphere angular velocity along Trajectory~1}
	\label{fig:angular_velocity_error}
\end{figure}

As shown in Fig.~\ref{fig:angular_velocity_error}, the maximum error on the sphere angular velocity along Trajectory~1 is about 30~deg/s. It should be noted that the root-mean square error of the angular velocity of the sphere is smaller about~$x_0$ than about~$z_0$. Although errors are present between the desired and obtained angular velocities of the end-effector, those experiments confirm the fidelity of the PSW kinematic model obtained in Eq.~(\ref{eq:W_SW}), the kinematic Jacobian matrix of the cable-actuated PSW being the opposite of the transpose of the wrench matrix~${\bf W}_{SW}$. The errors between measured and desired end-effector angular velocities are mainly due to the slippage between the omni-wheels and the sphere.

The experimental motions of the wrist along Trajectory~2 are shown in the \href{https://bit.ly/3hvpsL2}{video}\footnote[1]{\url{https://bit.ly/3hvpsL2}} from 1min~7s to 3min~41s. Let~$\theta_e$ denote the rotation angle of the top plate. $\theta_e$ is defined from~$^0\mathbf{R}_1$ as follows:
\begin{equation}\label{eq:thetae}
	\theta_e = \mathrm{arccos}\left(\frac{\mathrm{tr}(^0\mathbf{R}_1) - 1}{2}\right) \, , \, 0 \leq \theta_e \leq \pi 
\end{equation}

\begin{figure}[!htbp]
	\centering
	\includegraphics[width=1\linewidth]{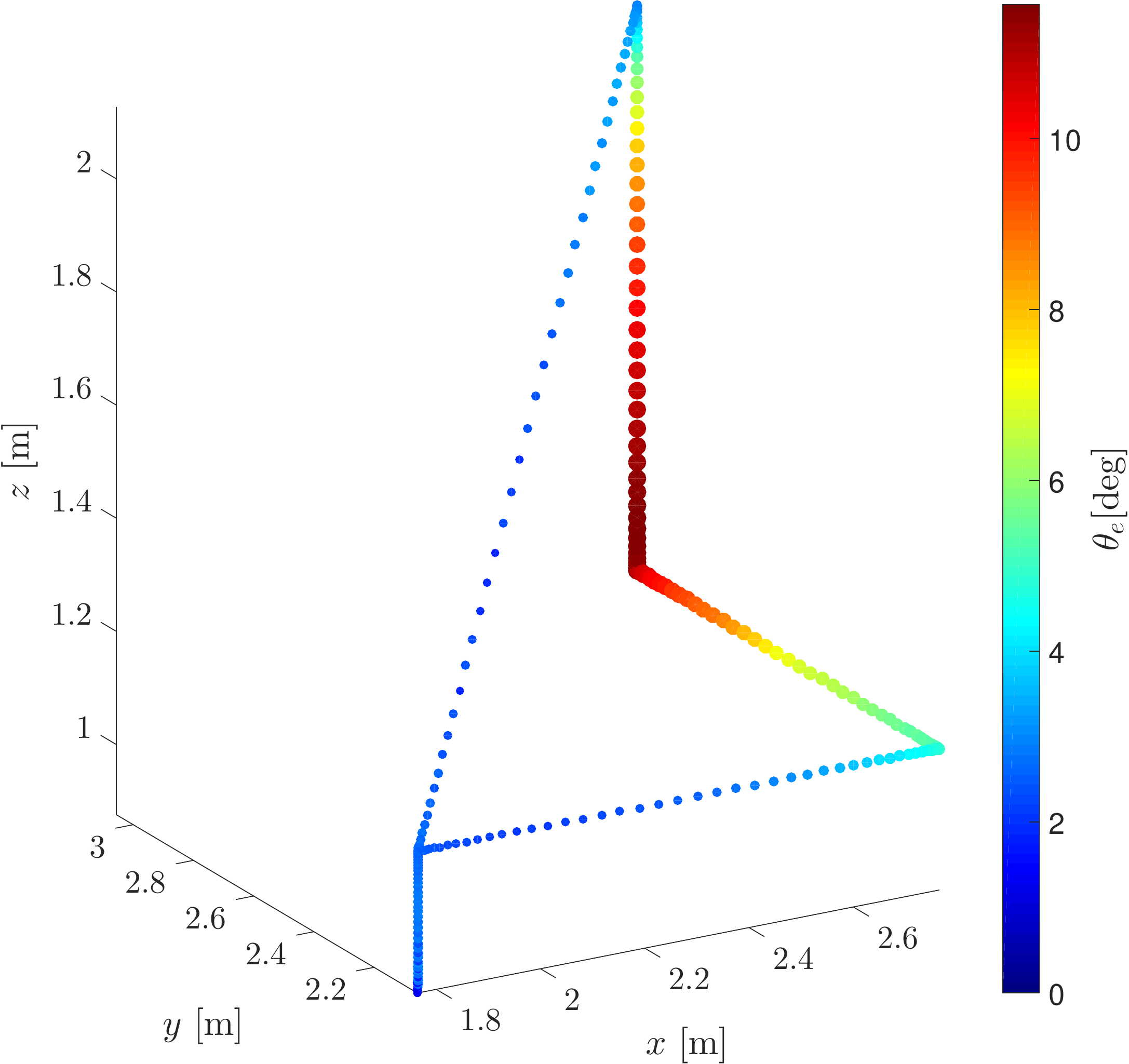}
	\caption{Parasitic inclination of the top plate along Trajectory~2}
	\label{fig:NU_R_translation_3D}
\end{figure}	

Figure~\ref{fig:NU_R_translation_3D} shows the evolution of~$\theta_e$, i.e., the parasitic inclination of the top plate, along the four straight line segments of Trajectory~2. Figure~\ref{fig:NU_R_translation_3D} illustrates the large translation workspace of the manipulator as presented in Figure \ref{subfig:workspace}. It is apparent that the maximum parasitic inclination angle of the top plate along this trajectory is about 10~deg. Furthermore, Fig.~\ref{fig:NU_R_translation_3D} shows the validity of the kinematic model of the top plate defined in Eq.~(\ref{eq:W_TP}).
The kinematic model of the robot is well validated quantitatively considering the small value of the parasitic inclinations. Nevertheless, its static model could not be validated other than by observing that inclinations of the top plate are mainly due to friction in the PSW to be identified and errors in the identified Cartesian coordinates of the cable exit points.

The experimental motions of the wrist along Trajectory~3 are shown in the \href{https://metillon.net/3T3R_JMR}{video}\footnotemark[1] from 3min~41s to 5min~38s. It is noteworthy that parasitic inclinations of the top plate appear once the sphere starts rotating. These parasitic motions are caused by several factors. Some improvements in the PSW design and robot control are needed to address these weaknesses. First, the cable anchor points located on the top plate and shown in Fig.~\ref{fig:anchor_points} are very close to each other leading to a high sensitivity of the orientation of the top plate to variations in cable lengths. Thus, this kinematic sensitivity should be reduced by moving the cable anchor points away from each other on the top plate. Besides, it is difficult to manage ($i$)~the friction between the sphere and the three bearings that support it, ($ii$)~the friction and contact between the omni-wheels and the sphere, and ($iii$)~the friction and slippage between the cable loops and the three pulleys around which the cables are wrapped. Indeed, it can be noted in the previous videos that the omni-wheels sometimes start to slide on the sphere and that they do not turn at other times when the cable loop is supposed to circulate. To resolve these issues, im t is necessary to better manage the cable tensions and improve the manufacturing and assembly of some parts such as the connection between the double omni-wheels and the top plate. Finally, the flexibility of the three arms holding the sphere may be responsible for bad contacts between the sphere and the omni-wheels.

\section{Conclusions}\label{sec:CONCLUSION}

This paper dealt with the design, modelling and prototyping of a hybrid robot. This robot, which is composed of a CDPR mounted in series with a PSW, has both a large translational workspace and an unlimited orientation workspace. It should be noted that the six degrees of freedom motions of the moving platform of the CDPR, namely, the base of the PSW, and the three-DOF motion of the PSW are actuated by means of eight actuators fixed to the base. As a consequence, the overall system is underactuated and its total mass and inertia in motion is reduced.

The kinetostatic model of the hybrid robot was expressed in order to determine its static equilibrium condition and find the cable arrangement maximizing its static workspace. A prototype of the CDPR with full-circle end-effector rotations was realized. The performances of the prototype were experimentally evaluated by measuring its angular-velocity response over and parasitic inclinations along three test trajectories. It appears that the maximum angular velocity error of the sphere is about 30~deg/s along the first trajectory for which the top plate remains on its support. Future work will focus on reducing parasitic inclinations of the end-effector thanks to an improved design of the wrist and better management of the cable tensions.

	

\begin{acknowledgment}
This work was supported by both the ANR CRAFT project, grant ANR-18-CE10-0004 and the RFI AtlanSTIC2020 CREATOR project. Assistance provided by Dr.~Saman Lessanibahri and M.~Julian Erskine through the experimentation process is highly appreciated.
\end{acknowledgment}

\begin{nomenclature}
	\entry{$\mathcal{F}_0(O_0,x_0,y_0,z_0)$}{Base frame attached to the rigid frame of the robot}
	\entry{$\mathcal{F}_1(O_1,x_1,y_1,z_1)$}{Frame attached to the top plate}
	\entry{$\mathcal{F}_2(O_2,x_2,y_2,z_2)$}{Frame attached to the end-effector of the Parallel Spherical Wrist}
	\entry{$\mathbf{l}_{i}$}{$i$-th cable vector pointing from $B_i$ to $A_i$, $i \in [\![1,\dots,8]\!]$}
	\entry{$\mathbf{a}_{i}$}{Cartesian coordinates vector of point $A_i$, $i \in [\![1,\dots,8]\!]$}
	\entry{$\mathbf{b}_{i}$}{Cartesian coordinates vector of point $B_i$, $i \in [\![1,\dots,8]\!]$}
	\entry{$\mathbf{p}$}{Cartesian coordinates vector of point $P$}
	\entry{$\mathbf{u}_{i}$}{$i$-th cable unit vector, $i \in [\![1,\dots,8]\!]$}
	\entry{$\mathbf{t}$}{Cable tension vector}
	\entry{$\mathbf{w}_{e}$}{External wrench vector}
	\entry{$\mathbf{w}_g^{TP}$}{External wrench vector associated to the top plate}
	\entry{$\mathbf{w}_g^{SW}$}{External wrench vector associated to the Parallel Spherical Wrist}
	\entry{$\mathbf{d}_i$}{Cross-product of $\mathbf{b}_i$ and $\mathbf{u}_i$, $i \in [\![1,\dots,8]\!]$ }
	\entry{$\mathbf{f}_{TP}$}{Force applied on the top plate}
	\entry{$\mathbf{m}_{TP}$}{Moment applied on the top plate}
	\entry{$\mathbf{m}_{SW}$}{External moments applied by the environment onto the end-effector}
	\entry{$\mathbf{v}_i$}{Axis of actuation force produced by the $i$-th omni-wheel on the sphere}
	\entry{$\mathbf{n}_i$}{Unit normal vector of $i$-th omni-wheel on plane $\pi_i$ at point $C_i$}
	\entry{$\mathbf{q}_{SW}$}{Orientation vector of the end-effector}
	
	\entry{$l_{i}$}{$i$-th cable length, $i \in [\![1,\dots,8]\!]$}
	\entry{$\textit{r}_s$}{Radius of the sphere}
	\entry{$\textit{r}_o$}{Radius of the omni-wheels}
	\entry{$r_d$}{Radius of the embedded drum of the cable loop}
	
	\entry{$A_i$}{$i$-th cable exit point, $i \in [\![1,\dots,8]\!]$}
	\entry{$B_i$}{$i$-th cable anchor point, $i \in [\![1,\dots,8]\!]$}
	\entry{$C_i$}{Contact point between the $i$-th omni-wheel and the sphere, $i \in [\![1,2,3]\!]$}
	\entry{$R_i$}{Available anchor point on the top plate $i \in [\![1,\dots,15]\!]$}
	
	\entry{$\dot{\varphi}_i$}{Angular velocity of the $i$-th omni-wheel, $i \in [\![1,2,3]\!]$}
	\entry{$\omega$}{3-dimensional angular velocity of the end-effector}
	
	\entry{$\alpha$}{Angle associated to the position of the contact points $C_i$ on the sphere, $\alpha \in [0,\pi]$}
	\entry{$\beta$}{Angle between tangent to the sphere and the actuation force of the omni-wheel, $\beta \in [-\frac{\pi}{2},\frac{\pi}{2}]$}
	\entry{$\gamma_i$}{Angle between the contact points $C_i$ and $\vec{x}_1$, $i \in [\![1,2,3]\!]$}
	\entry{$\theta$}{Pitch angle of the end-effector}
	\entry{$\psi$}{Roll angle of the end-effector}
	\entry{$\chi$}{Yaw angle of the end-effector}
			
	\entry{$\tau_i$}{$i$-th omni-wheel torque}
	
	\entry{${^0}\mathbf{R}_{1}$}{Rotation matrix from frame $\mathcal{F}_0$ to frame $\mathcal{F}_1$}
	\entry{$\mathbf{W}$}{Wrench matrix}
	\entry{$\mathbf{W}_{TP}$}{Wrench matrix related to the top plate}
	\entry{$\mathbf{W}_{SW}$}{Wrench matrix related to the Parallel Spherical Wrist}
	\entry{$\mathbf{A}$}{Forward Jacobian matrix of the Parallel Spherical Wrist}
	\entry{$\mathbf{B}$}{Inverse Jacobian matrix of the Parallel Spherical Wrist}
	\entry{$\mathbf{W}_{c}$}{Wrench matrix of the cable-loop}

	\entry{$\Pi_i$}{Plane passing through the contact point $C_i$ and tangent to the sphere, $i \in [\![1,2,3]\!]$}
	\entry{$\mathcal{L}_i$}{Line tangent to the sphere on $C_i$ in the plane $(\vec{x_1}, \vec{y_1})$}
	\entry{$\mathcal{S}$}{Static workspace of the manipulator}
	\entry{$\mathcal{S}_{AO}$}{Static workspace of the manipulator with a given orientation of the top plate}
	\entry{$\mathcal{R}_{\mathcal{S}}$}{Proportion of the static workspace to the overall space occupied by the manipulator}
	
	\entry{$N_{\mathcal{S}}$}{Number of points inside the discretized static workspace}
	
	\entry{$\mathcal{N}_e$}{Number of exit point combinations}
	\entry{$\mathcal{N}_a$}{Number of anchor points}
	\entry{$\mathcal{N}_C$}{Number of cable configurations}
	\entry{$\mathcal{N}_{CL}$}{Number of cable configurations considering cable-loop}
	\entry{$\textit{n}_e$}{Number of available exit-points}
	\entry{$\textit{n}_c$}{Number of cables}
	\entry{$\textit{n}_a$}{Number of selected anchor points}
	\entry{$n_{aSC}$}{Number of available anchor points for the single cable}
	\entry{$n_{aCL}$}{Number of available anchor points for cable-loop}
	
	\entry{$\mathcal{S}_C$}{Set of possible cable configurations}

\end{nomenclature}
	

%
	
\end{document}